\documentclass[a4paper,conference]{IEEEtran}
\usepackage[utf8]{inputenc}
\usepackage[T1]{fontenc} 
\IEEEoverridecommandlockouts

\usepackage{cite}
\usepackage{amsmath,amssymb,amsfonts}
\usepackage{algorithmic}
\usepackage{graphicx}
\usepackage{textcomp}
\usepackage{xcolor}
\usepackage[english]{babel}
\usepackage[letterpaper,top=2cm,bottom=2cm,left=2cm,right=2cm,marginparwidth=1.75cm]{geometry}
\usepackage[subpreambles=true]{standalone}
\usepackage[labelfont=footnotesize, textfont=footnotesize, figurename=Fig.]{caption} 
\usepackage{textcomp}
\usepackage{longtable}
\usepackage{flushend}
\usepackage[nohyperlinks, nolist]{acronym}
\def\BibTeX{{\rm B\kern-.05em{\sc i\kern-.025em b}\kern-.08em
    T\kern-.1667em\lower.7ex\hbox{E}\kern-.125emX}}

\makeatletter 
\newcommand{\linebreakand}{%
  \end{@IEEEauthorhalign}
  \hfill\mbox{}\par
  \mbox{}\hfill\begin{@IEEEauthorhalign}
}
\makeatother 

\usepackage{booktabs}
\usepackage{float} 

\usepackage{subfig} 

\usepackage{siunitx}

\usepackage{tikz}
\usepackage{wasysym}
\usepackage{makecell}
\usepackage{adjustbox}
\usepackage{tikz-3dplot}
\usetikzlibrary{automata}
\usetikzlibrary{positioning}
\usetikzlibrary{calc}
\usetikzlibrary{fit}
\usetikzlibrary{spy}
\usetikzlibrary{arrows.meta}
\usetikzlibrary{backgrounds}
\usetikzlibrary{plotmarks}
\usetikzlibrary{shapes,shapes.geometric,shapes.symbols,shapes.arrows, shapes.multipart}
\usetikzlibrary{patterns}

\begin{document}

\title{
Hybrid
Spiking Neural Networks 
for 
Low-Power 
Intra-Cortical Brain-Machine Interfaces

\thanks{
The project on which this report is based was sponsored by the German Federal Ministry of Education and Research under grant number 16ME0801. The responsibility for the content of this publication lies with the author.
}
}

\author{




}

\author{\IEEEauthorblockN{Alexandru Vasilache$^*$,\!$^{1,2}$ Jann Krausse$^*$,\!$^{1,3}$  \\ Klaus Knobloch,\!$^{3}$ Juergen Becker \!$^{1}$ \\
    }
    \IEEEauthorblockA{
    \textit{$^{1}$ Karlsruhe Institute of Technology, Karlsruhe, Germany}\\
    \textit{$^{2}$ FZI Research Center for Information Technology, Karlsruhe, Germany} \\
    \textit{$^{3}$ Infineon Technologies, Dresden, Germany} \\
    }
}

\maketitle
\def\thefootnote{*}\footnotetext{These authors contributed equally to this work.}\def\thefootnote{\arabic{footnote}}

\begin{acronym}[]
    \acro{ann}[ANN]{artificial neural network}
    \acro{snn}[SNN]{spiking neural network}
    \acro{ai}[AI]{artificial intelligence}
    \acro{ibmi}[iBMI]{intra-cortical brain-machine interface}
    \acro{bmi}[BMI]{brain machine interface}
    \acro{gru}[GRU]{gated recurrent unit}
    \acro{lif}[LIF]{leaky integrate-and-fire}
    \acro{sgru}[sGRU]{spiking GRU}
\end{acronym}

\begin{abstract}
\textbf{\Acp{ibmi} have the potential to dramatically improve the lives of people with paraplegia by restoring their ability to perform daily activities. 
However, current \acp{ibmi} suffer from scalability and mobility limitations due to bulky hardware and wiring. 
Wireless \acp{ibmi} offer a solution but are constrained by a limited data rate.
To overcome this challenge, we are investigating hybrid spiking neural networks for embedded neural decoding in wireless \acp{ibmi}. The networks consist of a temporal convolution-based compression followed by recurrent processing and a final interpolation back to the original sequence length. As recurrent units, we explore \acp{gru}, \ac{lif} neurons, and a combination of both  - \acp{sgru} and analyze their differences in terms of accuracy, footprint, and activation sparsity.
To that end, we train decoders on the "Nonhuman Primate Reaching with Multichannel Sensorimotor Cortex Electrophysiology" dataset and evaluate it using the NeuroBench framework, targeting both tracks of the IEEE BioCAS Grand Challenge on Neural Decoding.
Our approach achieves high accuracy in predicting velocities of primate reaching movements from multichannel primary motor cortex recordings while maintaining a low number of synaptic operations, surpassing the current baseline models in the NeuroBench framework.
This work highlights the potential of hybrid neural networks to facilitate wireless \acp{ibmi} with high decoding precision and a substantial increase in the number of monitored neurons, paving the way toward more advanced neuroprosthetic technologies.
}
\end{abstract}
\section{Introduction}
Tens of millions of lives worldwide are suffering from paralysis \cite{paralysis_us, spinalcordinjury_who}. Those affected experience an impaired ability to direct their movements, which, in severe cases, leads to a complete loss of motor control. This motivates the development of technology that can decode patients' brain activity and accordingly control assistive prostheses. Such devices are called \acp{bmi} \cite{bmi} and have been very successful with restoring motor control \cite{bmi_mood}, sensory information \cite{bmi_appl}, or even emotional responses \cite{bmi_mood}.

Usually, \acp{bmi} are directly placed on the surface of a patient's brain to ensure the maximal quality of the recorded brain signals (\acp{ibmi}). However, this raises two problems. First, implants are connected via bulky wiring to the operating equipment, severely restricting the patient's movement \cite{wireless_ibmi}. Second, permanently opening the skull to allow wiring increases the risk of infection \cite{infection_risk}. In hopes of mitigating this, research is 
moving towards wireless \acp{ibmi} \cite{wireless_ibmi, wireless_ibmi2}.

The Grand Challenge on Neural Decoding for Motor Control of non-Human Primates of IEEE BioCAS 2024 calls for solutions to the scalability issues of such wireless \acp{bmi}. Since data rates are limited due to bit-error rates, heat dissipation, and battery lifetime, an optimal solution should handle the trade-off between high-quality neural decoding, data compression, and resource management. As the development of techniques for embedded artificial intelligence progresses, neural networks present promising candidates for wireless low-power neural decoders \cite{ai_bmi, jann_socc}. Additionally, biologically inspired \acp{snn} benefit from high temporal sparsity, single-bit communication facilitated by spikes, and an intrinsic recurrence due to their statefulness \cite{snnnewdnn}. Consequently, participants of the Grand Challenge on Neural Decoding are tasked with training a neural network on the Primate Reaching dataset \cite{primate_reaching_dataset} for predicting the velocities of cursor movements. The network is then evaluated using the NeuroBench framework to obtain metrics regarding accuracy and resources \cite{neurobench}. Results are judged based on two challenge tracks: track 1 assesses sole accuracy optimization, while track 2 targets the co-optimization of accuracy, memory footprint, and number of compute operations, as defined in \cite{neurobench}.

Our work presents a hybrid network architecture 
of temporal convolutions in combination with recurrent processing and a subsequent interpolation back to the original sequence length. While \acp{gru} are very effective in sequence modeling \cite{gru}, networks based on spiking neurons like the \ac{lif} model profit from the advantages of \acp{snn} regarding resourcefulness mentioned above \cite{lif}. Hence, we investigate recurrent processing by \acp{gru}, \ac{lif} units, and a combination of both and discuss the differences in their results.

Furthermore, we motivate the chosen architecture via a few experiments before presenting the results of all three types of recurrence. 
All three network types beat the baselines given by \cite{neurobench} in at least one of the challenge tracks by a good margin. However, the different recurrence types show evident differences in accuracy and resourcefulness. Based on that, we will discuss the implications of using spiking elements. Finally, we point out the possibilities of the real-time deployment of these networks and areas of future work.
\section{Related Work}
The authors of \cite{rel_work_dethier} used \acp{snn} to predict a rhesus monkey's arm velocity accurately. However, the network was not trained directly on the data. Instead, they mapped a Kalman filter onto the network.

In \cite{rel_work_liao}, the authors train \acp{snn} on two datasets for offline finger velocity decodings. They achieve high accuracy and compare their approach to the \acp{ann} baseline, even specifying numbers for total operations and memory accesses. Still, their network represents a simple feed-forward architecture and is trained on a different dataset than this work.

The clear baseline for this work is given by \cite{neurobench}. Among other datasets, the authors make the dataset of \cite{primate_reaching_dataset} available for deep learning approaches and subsequently train neural networks as baselines. They differentiate between \acp{ann} and \acp{snn} as well as between networks that target pure reconstruction accuracy (track 1) and those that co-optimize accuracy and resource demands (track 2). The used networks are of relatively simple architecture. Their work aims to enable others to benchmark respective datasets easily. We will make use of their work and surpass their baseline using a different network architecture in both challenge tracks.
\section{Methods}
\subsection{Motivating an Interpolation-based Approach}
Our interpolation approach is inspired by observing primate cursor movements. In the video, a new target appears each time the previous one is reached, prompting a rapid, goal-directed movement toward it. This suggests that the movement can be approximated by discrete, target-locked actions rather than fine-grained continuous adjustments.

\begin{figure}[t]
    \centering
    \begin{tikzpicture}

\begin{axis} [
            name=interpol_demo,
            height=4.62cm, 
            width=10cm, 
            xmin=-0.5,
            xmax=0.7, 
            ymin=-0.25,
            ymax=0.25, 
            xlabel=$v_x$ / \si{\metre\per\second},
            ylabel=$v_y$ / \si{\metre\per\second},
            xtick={-0.4, -0.2, 0, 0.2, 0.4, 0.6},
            ytick={-0.2, 0, 0.2},
            xticklabels={-0.1, -0.05, 0, 0.05, 0.1, 0.15},
            yticklabels={-0.05, 0, 0.05},
            legend style={nodes={scale=0.7, transform shape}, at={(0.8,0.95)},anchor=north}
            ]
\addplot[darkgray, style=semithick] table {data/original_data_sample_0_copy.tex};
    
\addplot[teal, densely dashed, style=thick, opacity=0.7] table {data/interpolated_data_sample_0_copy.tex};

\addplot[only marks, teal, mark size=1.3, mark=*, fill opacity=0.3] table {data/interpolated_data_sample_0_keypoints.tex};
    
\legend{
original data, 
interpolated data,
keypoints
}
\end{axis}
 
\end{tikzpicture}
    \caption{Linear interpolation of discretized cursor velocities (8 steps) visualized above original cursor velocities.}
    \label{fig:label_interpolation}
\end{figure}
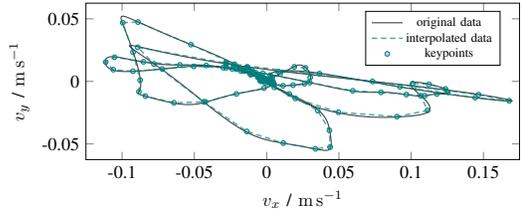

Based on this, we hypothesize that capturing a few keypoints along the velocity trajectory and interpolating between them can effectively approximate the whole movement velocity. Fig. \ref{fig:label_interpolation} illustrates this concept by comparing the original movement with a simplified version, where the velocity at every eighth point is retained, and linear interpolation is used between them. We argue that the resulting error is negligible, assuming that the keypoint prediction is of high quality, as the R2 score between the interpolated test set and the original test set is 0.998 with 4-step interpolation, 0.988 with 8-step interpolation and 0.955 with 16-step interpolation.

%



\subsection{Model Architecture}
\label{sec:model_architecture}

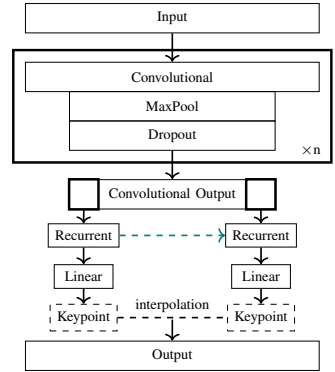
\begin{figure}[t]
   \centering
   \resizebox{0.49\textwidth}{!}{
\tikzstyle{every node}=[font=\scriptsize]
\begin{tikzpicture}[node distance=1cm]
    \node (input) [rectangle, draw, minimum width=5cm, minimum height=0.5cm] {Input};

    \node (conv1) [rectangle, draw, minimum width=5cm, minimum height=0.5cm, below of=input] {Convolutional};
    \node (pool1) [rectangle, draw, minimum width=3.5cm, minimum height=0.5cm, below of=conv1, yshift=0.5cm] {MaxPool};
    \node (drop1) [rectangle, draw, minimum width=3.5cm, minimum height=0.5cm, below of=pool1, yshift=0.5cm] {Dropout};
    \node (label_conv1) [right=0.3cm of drop1, yshift=-0.25cm] {$\times$n};
    \draw [black, very thick] ($(conv1.north west)+(-0.2,0.2)$)  rectangle ($(drop1.south east)+(0.95,-0.2)$); 


    \node (conv_out) [rectangle, draw, minimum width=3.5cm, minimum height=0.5cm, below of=drop1] {Convolutional Output};

    \node (gru1) [rectangle, draw, below of=conv_out, xshift=-1.51cm, minimum width=1cm, yshift=0.3cm] {Recurrent};
    \node (gru2) [rectangle, draw, below of=conv_out, xshift=1.51cm, minimum width=1cm, yshift=0.3cm] {Recurrent};
    \node (fc1) [rectangle, draw, below of=gru1, minimum width=1cm, yshift=0.3cm] {Linear};
    \node (fc2) [rectangle, draw, below of=gru2, minimum width=1cm, yshift=0.3cm] {Linear};
    \node (out1) [rectangle, draw, dashed, below of=fc1, minimum width=1cm, yshift=0.3cm] {Keypoint};
    \node (out2) [rectangle, draw, dashed, below of=fc2, minimum width=1cm, yshift=0.3cm] {Keypoint};
    \node (interp) [below of=conv_out, xshift=0cm, yshift=-0.9cm] {interpolation};
    \node (output) [rectangle, draw, minimum width=5cm, minimum height=0.5cm, below of=interp, yshift=0.15cm] {Output};

    \draw [->, thick] (input) -- (conv1);
    \draw [->, thick] (drop1) -- (conv_out);

    \draw [->, thick] ($(conv_out.south west)+(0.25,0)$) -- (gru1.north);
    \draw [->, thick] ($(conv_out.south east)+(-0.25,0)$) -- (gru2.north);
    \draw [->, thick, teal, dashed] (gru1) -- (gru2);
    \draw [->, thick] (gru1) -- (fc1);
    \draw [->, thick] (gru2) -- (fc2);
    \draw [->, thick] (fc1) -- (out1);
    \draw [->, thick] (fc2) -- (out2);
    \draw [thick, dashed] (out1) -- (out2);
    \draw [->, thick] (interp) -- (output);
    
    \draw [very thick] ($(conv_out.north west)+(0,0)$)  rectangle ($(conv_out.south west)+(0.5,0)$);
    \draw [very thick] ($(conv_out.north east)+(0,0)$)  rectangle ($(conv_out.south east)+(-0.5,0)$);
\end{tikzpicture}
}
   \caption{Illustration of the general network architecture used in this work.}
   \label{fig:net-arch}
\end{figure}

The general architecture of the model (Fig. \ref{fig:net-arch}) involves temporal convolutions to reduce the number of time steps in a sequence of neuron recordings from the input size of 1024 to the desired number of keypoints and efficiently extract temporal features. 
To create sufficient keypoint pairs, convolutional blocks reduce the sequence to a length of \textit{number of keypoints + 1}.
These features are then processed by recurrent units and a fully connected layer to determine output velocities as keypoints. We apply linear interpolation between the determined keypoints to scale the output sequence back to the original sequence length.

Here, we compare three types of recurrent units for the architecture described above. Those comprise \ac{gru} and \ac{lif} units, as well as a fusion of both, which we call the \ac{sgru}. We define the \ac{sgru} as

\begin{equation}
\mathbf{r}_t = \text{LIF}(\mathbf{W}_r \mathbf{x}_t + \mathbf{U}_r \mathbf{h}_{t-1}),
\end{equation}
\begin{equation}
\mathbf{z}_t = \text{LIF}(\mathbf{W}_z \mathbf{x}_t + \mathbf{U}_z \mathbf{h}_{t-1}),
\end{equation}
\begin{equation}
\mathbf{\tilde{h}}_t = \text{LIF}(\mathbf{W}_h \mathbf{x}_t + \mathbf{U}_h((1 - \mathbf{r}_t) \odot \mathbf{h}_{t-1})),
\end{equation}
\begin{equation}
\mathbf{h}_t = (1 - \mathbf{z}_t) \odot \mathbf{h}_{t-1} + \mathbf{z}_t \odot \mathbf{\tilde{h}}_t,
\end{equation}
where $\mathbf{r}_t$, $\mathbf{z}_t$, $\mathbf{\tilde{h}}_t$, and $\mathbf{h}_t$ denote reset gate, update gate, candidate hidden state, and hidden state at time $t$, respectively. $\mathbf{W}_r$, $\mathbf{W}_z$, $\mathbf{W}_h$, $\mathbf{U}_r$, $\mathbf{U}_z$, and $\mathbf{U}_h$ are learnable parameters. $\mathbf{x}_t$ denotes the input. LIF refers to the implementation of the \ac{lif} spiking neuron model presented in \cite{lif}.

\begin{figure}[t]
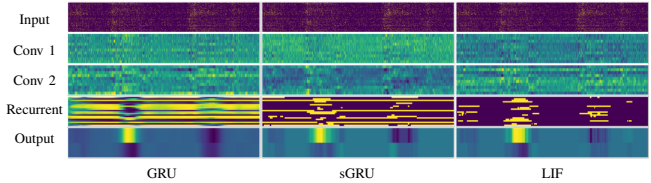

    \centering
    \includestandalone[width=1\linewidth]{figures/tikz/states_vis}
    \caption{Layer visualisations for \ac{gru}-t2, \ac{sgru}-t2, \ac{lif}-t2.}
    \label{fig:state-vis}
\end{figure}

Fig. \ref{fig:state-vis} displays the intermediary states of all three network types for visualization. Based on this general architecture, we present 2 model sizes, targeting track 1 (GRU-t1, sGRU-t1, LIF-t1) and track 2 (GRU-t2, sGRU-t2, LIF-t2) of the Neural Decoding Challenge.
The \ac{lif} networks additionally use recurrent weights.
Track 1 models employ three convolutional blocks with 32 channels, kernel sizes of 3, 6, and 12, and padding sizes of 5, 3, and 6, targeting 8-step interpolation with 127 keypoints. All max pooling layers use a kernel size and stride of 2. The size of the recurrent blocks is 64.
Track 2 models use two convolutional blocks with 10 channels and a kernel size of 3, which reduce the input size to 257 keypoints, effectively targeting a 4-step interpolation. To achieve the number of keypoints, the first convolutional layer uses a padding of 3, while the second convolutional layer employs a padding size of 1. The max pooling layers both use a kernel size and stride of 2. The size of the recurrent blocks is 20.
\section{Experiments and Observations}
To understand the relationship between model size and the R2 score and the tradeoff that comes with it, we trained four networks of different hidden sizes. Due to time limitations, we performed this experiment only for the \ac{sgru} model, training only on the indy2016062201 file with fewer data samples. Table \ref{tab:snn_size} displays the respective results.

\begin{table}[t]
\centering
\caption{Tradeoff between model size and R2 Score}
\label{tab:snn_size}
\resizebox{0.7\linewidth}{!}{
\begin{tabular}{cc}
\toprule
\textbf{Conv Channels x \ac{gru} Hidden Size} & \textbf{R2 Score} \\
\midrule
10x20 & 0.667 \\
32x64 & 0.692 \\
64x128 & 0.687 \\
128x256 & 0.661 \\
\bottomrule
\end{tabular}
}
\end{table}


Additionally, we study the influence of the number of keypoints on the R2 score by training four networks with 1025 to 129 keypoints (1-step to 8-step interpolation). Again, we trained the networks only on the indy2016062201 file with fewer data samples. Table \ref{tab:ann_keypoints} presents the results for the \ac{gru} model. Note that fewer keypoints directly translate to a higher R2 score. This trend was also confirmed for \ac{sgru}- and \ac{lif}-based networks.

\begin{table}[t]
\centering
\caption{Comparison of R2 scores for different number of keypoints. The networks are based on the \ac{gru} unit and do not differ in memory footprint.}
\resizebox{0.8\linewidth}{!}{
\label{tab:ann_keypoints}
\begin{tabular}{ccccc}
\toprule
\textbf{Keypoints (Interpolation)}  & \textbf{\makecell{Dense\\Operations}} & \textbf{\makecell{Effective\\MACs}} & \textbf{R2} \\
\midrule
1025 (1-step) & 46400.3 & 37184.3 & 0.736 \\
513 (2-step) & 27808.3 & 18592.3 & 0.766 \\
257 (4-step) & 20054.3 & 10838.3 & 0.764 \\
129 (8-step) & 17734.3 & 8518.3 & 0.779 \\
\bottomrule
\end{tabular}
}
\end{table}


We also ran experiments to evaluate the test performances of models trained on all three recordings for each primate. Interestingly, the R2 score decreases when using aggregated data, contrasting the expected increase in generalizability due to a more representative training set. This hints at a possible change or degradation of the signal recording from the intracortical electrodes across time.
\section{Results}
    



\subsection{Baseline Comparison}
We present the best results we obtained for \mbox{\ac{gru}-,} \mbox{\ac{sgru}-,} and \ac{lif}-based networks for challenge tracks 1 and 2 in Table \ref{tab:no_32_128}, on the metrics defined in \cite{neurobench}. Comparing our models to those provided by the baselines in \cite{neurobench}, we notice a larger footprint due to the increased input buffer size required for an input of size 1024 and the convolutional blocks. However, our models present fewer synaptic operations, judging by the Dense, MACs, and ACs values.

All our track 1 models achieve equal or higher R2 scores than the baselines, with \ac{gru}-t1 reaching an R2 score that is
increased from $0.615$ to $0.707$ compared to \mbox{B-ANN3}, while using 26\% fewer MACs and 34\% less Dense operations.

For track 2, we compare \ac{gru}-t2 with B-ANN2; it has only roughly 13\% of the MACs and an increased R2 score ($+0.045$). \ac{sgru}-t2 uses 60\% of the ACs, with the same R2 score when compared to B-SNN2 and 8\% of the MACs for the same R2 score when compared to B-ANN2. The \ac{lif}-t2 model achieves the same R2 score, with roughly the same activation sparsity, while only using 63\% of the Dense and  60\% of the ACs when compared to B-SNN2.

\subsection{Recurrence Comparison}
By far, the best performance has been achieved by the \ac{gru} recurrent unit for both investigated sizes. Furthermore, the \ac{gru} also gives the best trade-off between footprint and R2 score. Across both sizes, the lowest number of synaptic operations (Dense and MACs) is achieved by the \ac{lif} recurrence, which reaches the highest activation sparsity, as can be visually confirmed in Fig. \ref{fig:state-vis}. The \ac{sgru} recurrence achieves higher activation sparsity and lower MACs than the \ac{gru} for the same number of total synaptic operations and ACs at the cost of a higher footprint and lower R2 score. Compared to the \ac{lif}, \ac{sgru} consistently displays a slightly higher R2, hinting at improved memory management, compared to solely using \ac{lif} neurons.


\begin{table*}[ht]
\centering
\caption{
Results of the trained networks and their respective baselines. Networks prefixed with a B refer to the baselines given by \cite{neurobench}. Section \ref{sec:model_architecture} describes the corresponding network architectures. The exact definitions of each metric are defined in \cite{neurobench}. The values for Dense, MACs and ACs are computed by averaging the total over the length of the input (1024), as implemented by the Neurobench benchmarking tool \cite{neurobench}.
}
\label{tab:no_32_128}

\resizebox{0.8\linewidth}{!}{
\begin{tabular}{ccccccccc}
\toprule
\textbf{Track} & \textbf{Model} & \textbf{Footprint} & \textbf{\makecell{Connection\\Sparsity}} & \textbf{\makecell{Activation\\Sparsity}} & \textbf{Dense} & \textbf{MACs} & \textbf{ACs} & \textbf{R2} \\
\midrule

& B-ANN3 & 137752 & 0 & 0.681 & 33888 & 11507 & 0 & 0.615 \\
& B-SNN3 & \textbf{33996} & 0 & 0.788 & 43680 & 32256 & 5831 & 0.633 \\

\cmidrule(lr){2-9}

Track 1 & GRU-t1 & 352904 $\pm$ 0 & 0 $\pm$ 0 & 0 $\pm$ 0 & 22342 $\pm$ 0 & 8518 $\pm$ 0 & 793 $\pm$ 0 & \textbf{0.707 $\pm$ 0.012} \\

& sGRU-t1 & 425924 $\pm$ 0 & 0 $\pm$ 0 & 0.651 $\pm$ 0.017 & 22318 $\pm$ 0 & 7238 $\pm$ 24 & 797.7 $\pm$ 0.8 & 0.656 $\pm$ 0.013 \\

& LIF-t1 & 302492 $\pm$ 0 & 0 $\pm$ 0 & \textbf{0.939 $\pm$ 0.008} & \textbf{20766 $\pm$ 0} & 6414 $\pm$ 0 & 825 $\pm$ 4 & 0.648 $\pm$ 0.022 \\

\midrule
\midrule

& B-ANN2 & \textbf{27160} & 0 & 0.676 & 6237 & 4970 & 0 & 0.576 \\
& B-SNN2 & 29248 & 0 & \textbf{0.998} & 7300 & 0 & 414 & 0.581 \\

\cmidrule(lr){2-9}

Track 2 & GRU-t2 & 174104 $\pm$ 0 & 0 $\pm$ 0 & 0 $\pm$ 0 & 4947 $\pm$ 0 & 627 $\pm$ 0 & 248 $\pm$ 0 & \textbf{0.621 $\pm$ 0.014} \\

& sGRU-t2 & 180716 $\pm$ 0 & 0 $\pm$ 0 & 0.69 $\pm$ 0.07 & 4932 $\pm$ 0 & 379 $\pm$ 23 & 250.2 $\pm$ 0.8 & 0.577 $\pm$ 0.013 \\

& LIF-t2 & 168596 $\pm$ 0 & 0 $\pm$ 0 & 0.946 $\pm$ 0.009 & \textbf{4631 $\pm$ 0} & 201 $\pm$ 0 & 254 $\pm$ 0.8 & 0.566 $\pm$ 0.016 \\

\bottomrule
\end{tabular}
}
\end{table*}

\section{Discussion}


We hypothesize that the reason for the higher achieved R2 score, given a sufficiently large receptive field (as seen in our models proposed for track 1), may be that the filtering operations performed by the convolutional layers offer better information aggregation across time, compared to the simple summing aggregation used by the baseline model B-SNN3.
\begin{figure}[t!]
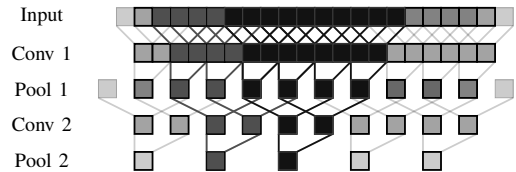

    \centering
    \includestandalone[width=0.8\linewidth]{figures/tikz/architecture_vis}
    \caption{Receptive Field Visualisation}
    \label{fig:rec-field}
\end{figure}

The proposed models use an input buffer window of 1024 steps provided by the NeuroBench \cite{neurobench} Primate Reaching Dataset, where each step represents 4 \si{\milli\second}. This results in a total buffer window and a latency of 4.096 \si{\second}. The models are executed for non-overlapping windows of size 1024, meaning that the model execution rate is 0.244 \si{\hertz}.

Our current approach comes with a high flexibility in the possible latency and execution rate that it can achieve, as both the convolutional and the recurrent layers allow for iterative data processing. Models \ac{gru}-t2, \ac{sgru}-t2, and \ac{lif}-t2 use a kernel size of 3 applied in two convolutional blocks. The receptive field determined by this structure can be visualized in Fig. \ref{fig:rec-field}. With the sizes mentioned above, the computation of one keypoint requires an effective buffer window of 10 steps, which offers a latency of 40 \si{\milli\second}. This would also reduce the input buffer size from 1024 to 10, reducing the model footprints by a sizable amount. The stride of the receptive field is 4 steps, or 16 \si{\milli\second}, which translates to an execution rate of 62.5 \si{\hertz}.
The theoretical upper limit of the latency of our models (40 \si{\milli\second}) is well under the time delay between stimulus and voluntary muscle movement reported by the neuroscience literature \cite{kurtzer_long-latency_2015}, which is typically greater than 100 \si{\milli\second}. Assuming no further latencies arise from signal transmission and ignoring computation time, our approach would be suitable for deployment in the real world, given an appropriate real-time implementation of the networks.

\begin{figure}[t]
    \centering
    \begin{tikzpicture}[spy using outlines={rectangle, gray, magnification=28,
                       size=3cm, connect spies}]
 
\begin{groupplot}[group style={group size=1 by 3, vertical sep=5pt, horizontal sep=5pt}, height=5cm, width=9.25cm]

    \nextgroupplot[
            ytick={-0.5, 0, 0.5},
            yticklabels={-0.125, 0, 0.125},  
            xticklabels=\empty,
            xmin=-1.6,  
            xmax=1.99,
            ymin=-0.6,  
            ymax=0.8,
            ]
    \addplot[darkgray] table {data/target_sample62.tex};

    \addplot[blue, densely dashed, style=thick] table {data/gru_sample62.tex};

    \addplot[cyan, densely dotted, style=thick] table {data/sgru_sample62.tex};

    \addplot[teal, densely dashdotted, style=thick] table {data/lif_sample62.tex};

    \node[] at ({-1.47, 0.7}) {\LARGE{\textbf{a}}};

    \nextgroupplot[
            ylabel shift=100pt,
            ytick={-0.5, 0, 0.5},
            yticklabels={-0.125, 0, 0.125},  
            xticklabels=\empty,
            xmin=-1.6,
            xmax=1.99,
            ymin=-0.6,
            ymax=0.8,
            ylabel=$v_y$ / \si{\metre\per\second},
            y label style={yshift=-105pt},
            legend style={nodes={scale=0.62, transform shape}, at={(0.88,0.95)},anchor=north},
            ]
    \addplot[darkgray] table {data/target_sample100.tex};

    \addplot[blue, densely dashed, style=thick] table {data/gru_sample100.tex};

    \addplot[cyan, densely dotted, style=thick] table {data/sgru_sample100.tex};

    \addplot[teal, densely dashdotted, style=thick] table {data/lif_sample100.tex};

    \node[] at ({-1.47, 0.7}) {\Large{\textbf{b}}};

    \legend{ 
        target,
        GRU,
        sGRU,
        LIF,
    }

    \nextgroupplot[
            ytick={-0.5, 0, 0.5},
            yticklabels={-0.125, 0, 0.125},  
            xtick={-1.5, -1, -0.5, 0, 0.5, 1, 1.5},
            xticklabels={-0.375, -0.25, -0.125, 0, 0.125, 0.25, 0.375},  
            xmin=-1.6,
            xmax=1.99,
            ymin=-0.6,
            ymax=0.8,
            xlabel=$v_x$ / \si{\metre\per\second},
            scaled ticks=base 10:0,
            ]
    \addplot[blue, mini dashed, line width=0.04pt] table {data/gru_sample53.tex};

    \addplot[cyan, line width=0.04pt, dots=400 per 1cm] table {data/sgru_sample53.tex};

    \addplot[teal, mini dashdotted, line width=0.04pt] table {data/lif_sample53.tex};

    \addplot[darkgray, line width=0.04pt] table {data/target_sample53.tex};

    \node[] at ({-1.47, 0.7}) {\LARGE{\textbf{c}}};

    \spy [] on (3.39, -5.74) in node at (5.9,-5.48);
    
\end{groupplot}
 
\end{tikzpicture}
    \caption{Visualization of the velocity outputs of all three model types for three exemplary samples each. \textbf{a} displays a sample learned well by all three networks (R2 $\approx$ 0.9). \textbf{b} shows the output for a sample for which the networks display average accuracy (R2 $\approx$ 0.7). For the sample shown by \textbf{c}, the networks could not accurately reconstruct the target (R2 $\ll$ 0).}
    \label{fig:models_output}
\end{figure}
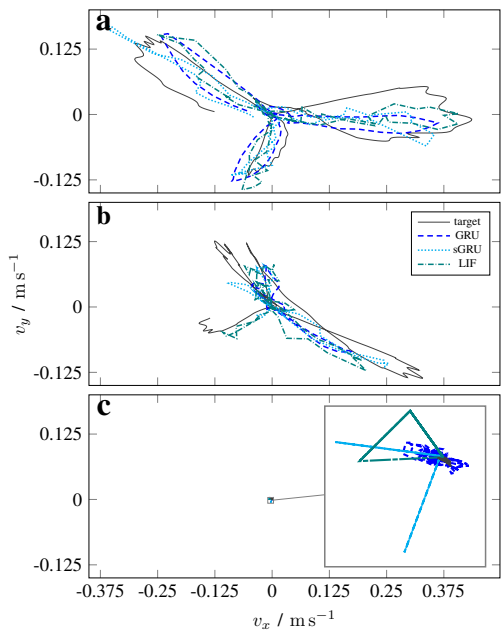
\section{Conclusion and Outlook}
\label{sec:conclusion}
This work targets both tracks of the Grand Challenge on Neural Decoding for Motor Control of non-Human Primates of IEEE BioCAS 2024. This includes track 1, which focuses on maximizing task accuracy, and track 2, which aims at co-optimizing accuracy and resource demand, which is critical for wireless \acp{ibmi}. The networks presented in this work surpass the baselines in \cite{neurobench} by good margins for both tracks.

For track 1, \ac{gru}- and \ac{sgru}-based networks beat the baselines by up to 7.4\% in terms of R2 while the \ac{lif}-based networks perform equal. For track 2, considering the margin of error, all networks are at least equal in R2 but show an improvement in the double-digit percentages in terms of compute operations. Only the footprint is increased by a rough factor of 6. We explain that this difference is due to large data buffers in our current model implementation. This gap could be eliminated in real-world deployment by taking advantage of the iterative nature of convolutional filters and recurrent units. Generally, the \ac{gru}-based networks score the highest in both tracks. However, the total amount of operations is the fewest for the \ac{lif}-based networks. Our \ac{sgru}-based models consistently achieve a higher R2 than solely using \ac{lif}-neurons, suggesting that such spiking neuron models could benefit from improved memory management.

Our work does not yet leverage some \ac{snn}-centered methods to improve their resourcefulness. This includes spike regularization, pruning, and event-triggered updating of the neural units, which will be included in future work. Finally, three of the six recordings in the dataset consist of motor cortex and somatosensory cortex recordings. We do not yet distinguish between the two different data types and expect an improved regression if done so.

Our work enhances the baseline for the primate reaching dataset and demonstrates the potential of using hybrid neural networks for efficient neural decoders. This advances the field of wireless \acp{ibmi} to eventually improve the lives of millions of humans suffering from paralysis.

\bibliographystyle{ieeetr}
\bibliography{refs}

\end{document}